\documentclass[10pt,twocolumn,letterpaper]{article}

\usepackage{iccv}
\usepackage{times}
\usepackage{epsfig}
\usepackage{graphicx}
\usepackage{amsmath}
\usepackage{amssymb}

\usepackage{booktabs}
\usepackage{multirow}
\usepackage{bm}
\usepackage{url}
\usepackage{svg}
\usepackage{comment}

\usepackage[pagebackref=true,breaklinks=true,colorlinks,bookmarks=false]{hyperref}

\usepackage[capitalize]{cleveref}
\crefname{section}{Sec.}{Secs.}
\Crefname{section}{Section}{Sections}
\Crefname{table}{Table}{Tables}
\crefname{table}{Tab.}{Tabs.}

\iccvfinalcopy % *** Uncomment this line for the final submission

\def\iccvPaperID{12423} % *** Enter the ICCV Paper ID here

\begin{document}

%%%%%%%%% TITLE
\title{INSTA-BNN: Binary Neural Network with INSTAnce-aware Threshold}

\begin{comment}
\end{comment}
\author{Changhun Lee$^{1}$ \qquad Hyungjun Kim$^{2}$ \qquad Eunhyeok Park$^{1}$ \qquad Jae-Joon Kim$^{3}$\\
\parbox{40em}{\small\centering $^{1}$ Pohang University of Science and Technology (POSTECH), Pohang, Korea}\\
\parbox{40em}{\small\centering $^{2}$ SqueezeBits Inc., Seoul, Korea \qquad $^{3}$ Seoul National University, Seoul, Korea}\\
}

\maketitle

%%%%%%%%% ABSTRACT
\begin{abstract}
Binary Neural Networks (BNNs) have emerged as a promising solution for reducing the memory footprint and compute costs of deep neural networks, but they suffer from quality degradation due to the lack of freedom as activations and weights are constrained to the binary values.
To compensate for the accuracy drop, we propose a novel BNN design called Binary Neural Network with INSTAnce-aware threshold (INSTA-BNN), which controls the quantization threshold dynamically in an input-dependent or instance-aware manner. According to our observation, higher-order statistics can be a representative metric to estimate the characteristics of the input distribution. INSTA-BNN is designed to adjust the threshold dynamically considering various information, including higher-order statistics, but it is also optimized judiciously to realize minimal overhead on a real device. 
Our extensive study shows that INSTA-BNN outperforms the baseline by 3.0\% and 2.8\% on the ImageNet classification task with comparable computing cost, achieving 68.5\% and 72.2\% \linebreak top-1 accuracy on ResNet-18 and MobileNetV1 based models, respectively.
\end{abstract}

%%%%%%%%% BODY TEXT
\section{Introduction}
\label{sec:intro}

Deep neural networks (DNNs) are well-known for their abilities across diverse vision tasks, \eg, image classification~\cite{he2016deep,krizhevsky2012imagenet,simonyan2014very,szegedy2015going}, object detection~\cite{lin2017focal,liu2016ssd,redmon2016you,ren2015faster}, and semantic segmentation~\cite{long2015fully,noh2015learning}. These DNNs usually achieve excellent accuracy by using a large model, but the large model having massive memory usage and computing cost prevents us from deploying it on mobile devices having insufficient resources. 
In order to minimize the computation and memory overhead, network binarization is an appealing optimization because weights and activations are quantized into 1-bit precision domain. Binary Neural Networks (BNNs) can achieve 32$\times$ reduction in memory requirement compared to their 32-bit floating point counterparts, and most modern CPUs or GPUs can serve the binary logic operations much faster than 32-bit floating point operations.

However, BNNs usually suffer from accuracy degradation due to the aggressive data quantization. 
In spite of the great efficiency of BNN, the accuracy degradation limits the deployment of BNNs in real-world applications.
Therefore, a large number of techniques have been introduced to minimize the accuracy degradation of BNNs focusing on weight binarization~\cite{qin2020forward,rastegari2016xnor}, shortcut connections~\cite{liu2018bi}, and threshold optimization~\cite{kim2021improving,liu2020reactnet,wang2020sparsity}.

\begin{figure}[t]
    \centering
    \includegraphics[width=0.95\linewidth]{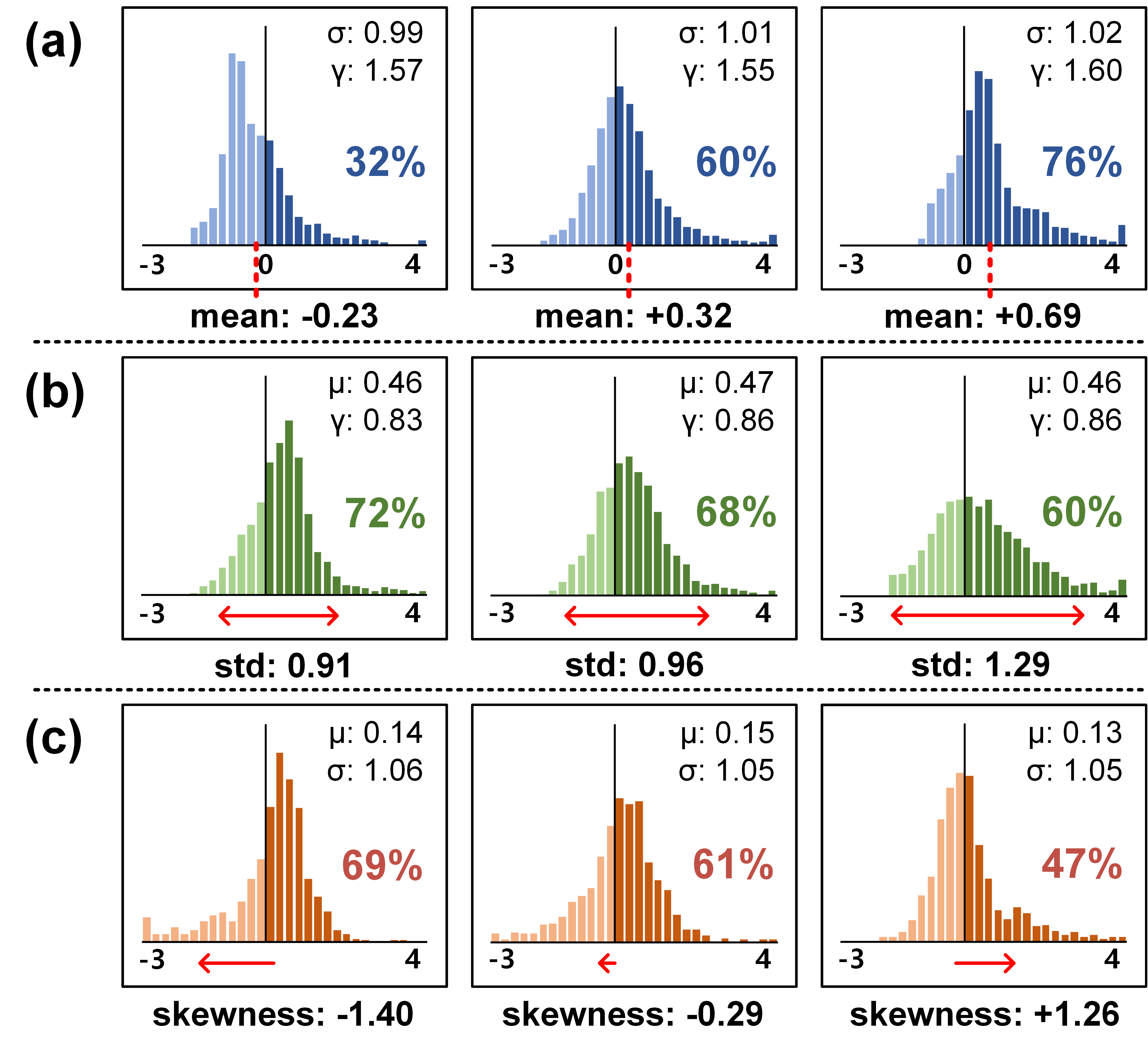}
    \caption{
    Instance-wise (a) mean, (b) standard deviation, and (c) skewness of pre-activation from three different instances.
    The percentage in each distribution represents the ratio of +1's in binary activation when sign function is used as activation function (threshold is 0). When observing one component (\eg mean ($\mu$) for the (a) case), the other two (\eg std ($\sigma$), skewness ($\gamma$)) are matched to be as close as possible. The other two values are displayed on the top right of each subgraph.}
    \label{fig:mean_std_skewness}
\end{figure}

In BNNs, only two values ($+1$ and $-1$) are available for activations.
Therefore, the threshold of quantization that decides the mapping to either $+1$ or $-1$ plays a critical role in BNNs.
Some of the previous works tried to train the thresholds of the binary activation function to control the activation distributions via back-propagation, where the threshold is fixed statically after the training~\cite{liu2020reactnet,wang2020sparsity}. However, one can easily observe that the per-instance activation is heavily distorted depending on the input data (\cref{fig:mean_std_skewness}), even while the batch-wise statistics are stabilized via batch normalization. As shown in the figure, the per-instance statistics numerically indicate the large distortion of pre-activation, and the corresponding outputs have a heavy fluctuation of the ratio of $+1$'s. In BNNs, static thresholds may not provide the adequate threshold for each instance, thereby resulting in sub-optimal results in terms of overall accuracy. 

In this work, we propose a novel INSTA-BNN that calculates the thresholds dynamically using the input-dependent or instance-wise information (\eg mean, variance, and skewness). The instance-aware threshold enriches the quality of binary features significantly, resulting in higher accuracy of BNNs. In addition, we provide a variant of Squeeze-and-Excitation (SE)~\cite{hu2018squeeze} with instance-wise adaptation as an additional option that one can exploit for even higher accuracy with extra parameters.
The proposed modules are extensively evaluated on a real device with a large-scale dataset~\cite{deng2009imagenet}, and we provide a practical guideline for network design that maximizes the benefit of INSTA-BNN while minimizing the increased overhead of it. With the aid of the guidelines, INSTA-BNN achieves higher accuracy with a similar number of parameters/operations and latency compared to previous works. We claim that the proposed INSTA-BNN can be an attractive option for BNN design.

%-------------------------------------------------------------------------
\section{Related works}
\label{sec:related_work}

\paragraph{Network Binarization}
Network binarization can significantly reduce the size of a model and the compute cost by reducing the precision of weights and activations to 1-bit. 
Hubara \etal \cite{hubara2016binarized} proposed to binarize both the weights and activations of a model, and suggested to use the straight-through-estimator (STE)~\cite{bengio2013estimating} for binarization function. 
XNOR-Net~\cite{rastegari2016xnor} introduced the real-valued scaling factor for binarized weights and activations to minimize the quantization error. 
To increase the representational capability of a model, a real-valued shortcut was suggested in Bi-real-net~\cite{liu2018bi}.
There are several more approaches to improve the accuracy of binarized networks, such as exploiting extra component~\cite{martinez2020training,wang2019learning}, 
new model structures~\cite{bulat2017binarized,liu2018bi}, 
and advanced training techniques~\cite{kim2019binaryduo,martinez2020training,qin2020forward}.
In most previous BNNs, binarization of weights and activations were conducted with the Sign function:
\begin{equation}
    x_b = Sign(x_r) =
    \begin{cases} 
    +1,  \quad if \: x_r \geq 0 \\ 
    -1,  \quad if \: x_r < 0
    \end{cases}.
    \label{eq:sign}
\end{equation}
In case of activations, \cref{eq:sign} is also called the binary activation function.
The $x_r$ indicates a real-valued pre-activation value, and $x_b$ is a binarized ($+1$ and $-1$) activation value.
The value $x_r$ is compared with the threshold of the binary activation function.

\paragraph{Threshold Optimization}
Several works proposed to modify the thresholds of binary activation functions to improve the accuracy of BNNs.
ReActNet~\cite{liu2020reactnet} proposed the activation function with learnable threshold for which channel-wise parameters were trained via back-propagation.  
Although the proposed technique improved the accuracy of BNNs, the learned thresholds were fixed after training so that the same thresholds were always used for every input.
Kim \etal\cite{kim2021improving} pointed out that back propagation alone cannot train the thresholds to have an optimal value.
They demonstrated that the distribution of learned thresholds strongly depended on the initial values and did not deviate from the initial point much.
While Kim \etal showed that initializing the thresholds with small positive values improved the accuracy of BNNs, they did not provide how to find optimal threshold values analytically.
In summary, previous methods which tried to find optimal thresholds are sub-optimal, and all of them used the fixed threshold values during inference, failing to consider instance-wise statistical information of activation distributions.

%------------------------------------------------------------------------
\section{Instance-Aware BNN}
\label{section:insta-bnn}

\subsection{Importance of instance-wise threshold}
\label{section:instance}
Compared to full precision DNNs, BNNs are more sensitive to the distribution of pre-activation values since the binarized activations are constrained to either $+1$ or $-1$.
\cref{fig:mean_std_skewness}a shows the distribution of pre-activations for three different images.
Note that all three distributions are from the same layer and channel of a model, and they are the outputs after the batch normalization layer.
As shown in the figure, the mean of each distribution has a non-zero value because, even with batch normalization, the mean value of activation depends on the distribution of the corresponding input instance. 
However, the small mean drift in the pre-activation results in a large difference of binarized activation.
In \cref{fig:mean_std_skewness}a, 
while the two distributions (left and center) have mean values that are relatively close to 0 ($-$0.23 and +0.32), the ratio of $+1$'s shows a large difference (32\% and 60\%).
The static threshold of BNNs, combined with the insufficient expression capability, is not enough to generate high-quality binary outputs adequately aware of the delicate change of input data. 

%-------------------------------------------------------------------------
\begin{figure}
    \centering
    \includegraphics[width=0.95\linewidth]{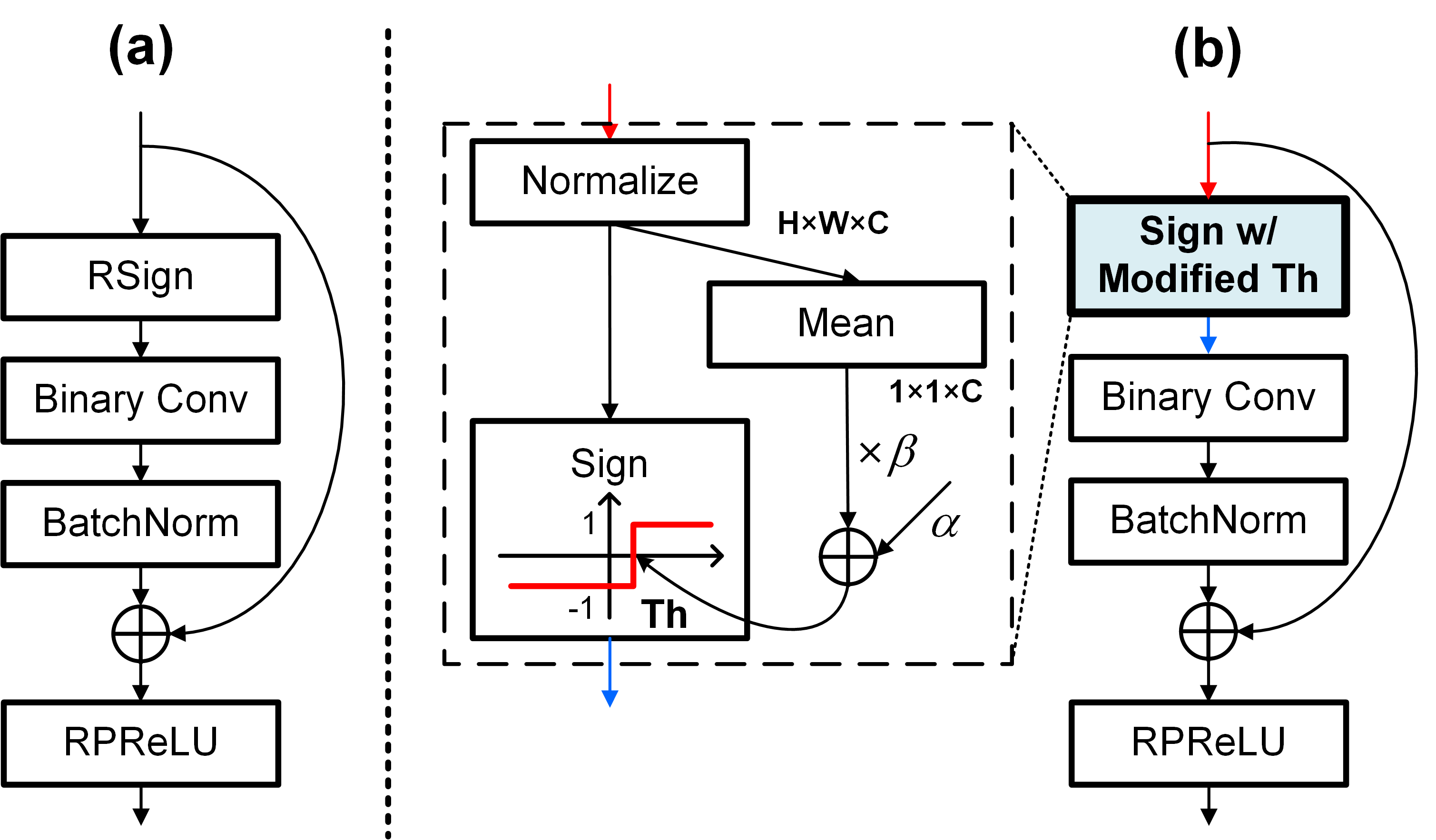}
    \caption{Convolution blocks and the proposed module structure that uses the instance-wise mean information. (a) Binary convolution block structure without instance-aware modules. (b) Binary convolution block that contains the proposed module.}
    \label{fig:structure_ori_and_mean}
\end{figure}

To address the aforementioned issue, we introduce a novel module that determines the quantization threshold considering the instance-wise information.
\cref{fig:structure_ori_and_mean}a shows the conventional block structure used in ReActNet~\cite{liu2020reactnet}.
To use the channel-wise learnable thresholds, ReActNet modified the Sign activation function as follows:
\begin{equation}
     x_b =
     \begin{cases} 
     +1,  \;\;\; if \;\; x_r \geq \alpha \\ 
     -1,  \;\;\; if \;\; x_r < \alpha
    \end{cases}.
 \label{eq:rsign}
\end{equation}
Note that all the learnable variables used in this section are channel-wise variables.

While the parameter $\alpha$ was trained using back-propagation in ReActNet, we propose to replace it with an instance-aware threshold ($TH$) as follows:
\begin{equation}
    x_b = 
    \begin{cases}
     +1, \;\;\; if \;\; x_r \geq TH\\
     -1, \;\;\; if \;\; x_r < TH
    \end{cases}.
    \label{eq:threshold}
\end{equation}
The instance-aware threshold is updated using per-instance statistical information, and we need to design an appropriate module for it. The module should be simplified as much as possible to minimize the computation/storage overhead, but it should be able to reflect the key information of input activation. 

The most straightforward implementation is extracting input data information and linearly combining it to update $TH$. We validate the effect of instance-wise $TH$ based on the mean of activation, which has a critical impact on the ratio of $+1$'s and $-1$'s, as shown in \cref{fig:mean_std_skewness}a. To maximize the benefit of $TH$ generation process, we utilize the difference between the batch-wise statistics and instance-wise statistics to see how much each instance deviates from the batch statistics. To calculate the drift between the per-instance mean and the mean over the entire dataset, we first normalize the pre-activation values using the Batch Normalization (BN) layer without applying the affine transformation. 
After the normalization layer, the instance-wise mean is calculated as follows:
\newcommand{\CLB}{\mkern -3.0mu \left[}
\newcommand{\CRB}{\right]}
\begin{equation}
    \Tilde{x}_r = \frac{x_r - \hat{\mu}}{\hat{\sigma}} \qquad  
    E\CLB\Tilde{x}_r\CRB = \frac{1}{H\times W}\sum_{i=1}^{H}\sum_{j=1}^{W}\Tilde{x}_r(i,j),
\end{equation}
where $H$ and $W$ are the height and width of the feature map. Using the instance-wise mean information, we formulate the instance-aware threshold as follows:
\begin{equation}
    TH = \alpha + \beta E\CLB\Tilde{x}_r\CRB.
    \label{eq:th_mean}
\end{equation}
The parameters $\alpha$ and $\beta$ are learned via back-propagation and jointly trained with model weights.
In addition to the learnable parameter $\alpha$ that is used in RSign (\cref{eq:rsign}) of ReActNet, we use an additional scaling factor $\beta$ which is related to instance-wise statistics. The proposed module is shown in \cref{fig:structure_ori_and_mean}b.

\begin{figure}[t]
    \centering
    \includegraphics[width=0.95\linewidth]{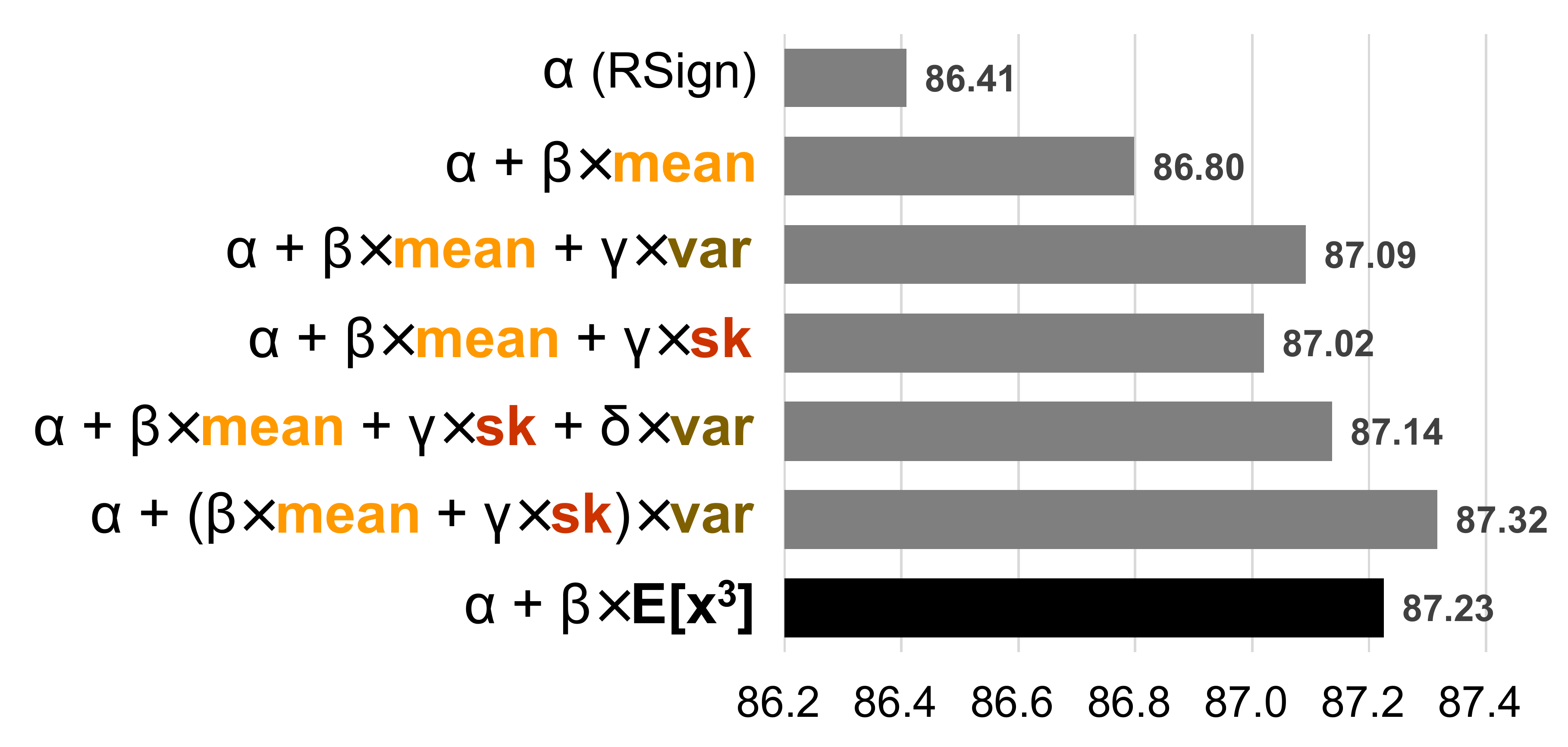}
    \caption{
    Test accuracy of binary ResNet-20 based model on CIFAR-10 dataset when different instance-wise statistic information are used for thresholds.
    Various combinations of mean, variance (var), and skewness (sk) are evaluated.
    }
    \label{fig:cifar10}
\end{figure}

We evaluate the effect of instance mean-aware thresholds on the ResNet-20 based model trained on the CIFAR-10 dataset. We compare the average accuracy of 6 runs with different random seeds.
As shown in the top two rows in \cref{fig:cifar10}, the proposed instance mean-aware threshold improves the accuracy by 0.39\% compared to the input-agnostic learnable threshold of RSign. 

\begin{figure*}[t]
    \centering
    \includegraphics[width=0.9\linewidth]{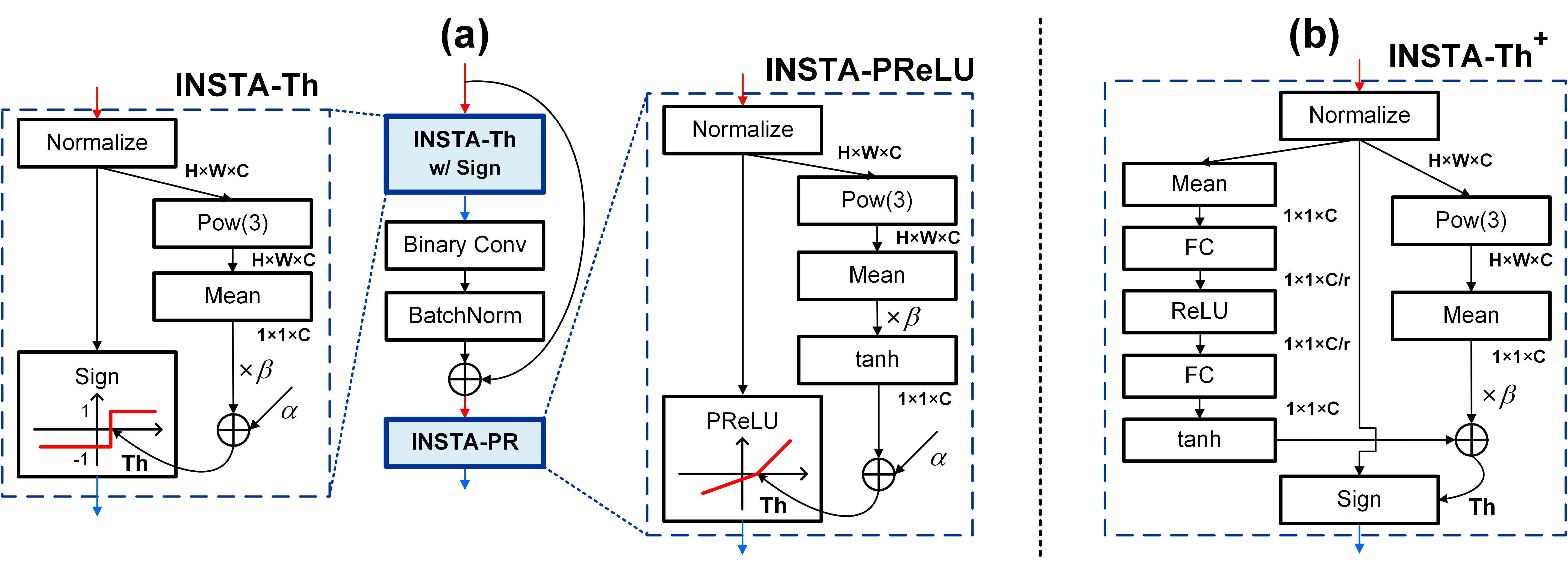}
    \caption{Convolution blocks and the proposed module structures of INSTA-BNN. (a) Binary convolution block that contains the proposed INSTA-Th and INSTA-PReLU modules. (b) INSTA-Th$^+$ module structure with SE-like block. INSTA-Th$^+$ can replace the INSTA-Th module in the structure of (a). Pow(3) indicates cubic operation, $x^3$.}
    \label{fig:structure}
    \vspace{-2mm}
\end{figure*}

%------------------------------------------------------------------------
\subsection{Importance of higher-order statistics information}
In the previous section, the simple instance-aware threshold works successfully with the mean of activation. However, we can easily identify that higher-order statistics, \eg variance and skewness of distribution, also well describe the important characteristics of activation (\cref{fig:mean_std_skewness}b and c). In this work, the skewness of data is estimated based on Fisher's moment coefficient of skewness~\cite{enwiki:1140693810}:
\begin{equation}
\begin{split}
  Skew&ness(X) =  \: E\CLB\left(\frac{X - \mu}{\sigma}\right)^3\CRB \\
           = & \: \frac{E\CLB X^3\CRB - 3\mu E\CLB X^2\CRB + 3\mu^2 E\CLB X\CRB - \mu^3}{\sigma^3} \\
          = & \: \frac{E\CLB X^3\CRB - 3\mu \sigma^2 - \mu^3}{\sigma^3}.
\end{split}
  \label{eq:skewness}
\end{equation}
As shown in the \cref{fig:mean_std_skewness}b and c, even when the mean values are comparable, the characteristics of the binary activation could be highly varied depending on the variance or skewness of data distribution. Especially the skewness of distribution has been neglected so far, but it contains extra information that is not captured by mean and variance values (\cref{fig:mean_std_skewness}c).
Therefore, instance-wise $TH$ module should consider higher-order statistics jointly.

In order to validate the importance of higher-order information, we modify $TH$ (\cref{eq:th_mean}) in a diverse way to add the influence of variance and skewness information on thresholds (\cref{fig:cifar10}).
As expected, adding more statistical information for the instance-aware threshold improves the accuracy even further. When $TH=\alpha+(\beta\times mean+\gamma\times skewness)\times var$, we could achieve 87.32\% accuracy which is about 0.9\% higher than the baseline that uses ReActNet's learnable threshold.

While using more per-instance statistical information improves accuracy, computing such information for each instance requires a high computing cost.
For example, calculating all of mean, variance, and skewness of pre-activation for each instance and each channel is highly costly, incurring long inference latency.
Here we introduce a simple alternative way to consider instance-wise mean, variance, and skewness at the same time.
First, we can re-write the \cref{eq:skewness} as follows:
\begin{equation}
E\CLB X^3\CRB = (Skewness(X))\sigma^3 + 3 \mu \sigma^2 + \mu^3 .
\label{eq:approx}
\end{equation}
We can see that the equation has information of mean, variance, and skewness and produces similar accuracy (87.23\%) to the case that achieves the highest accuracy, 87.32\% (\cref{fig:cifar10}).
Therefore, we propose to use higher-order $E\CLB X^3\CRB$ term instead of calculating all the mean, variance, and skewness of the pre-activation distribution separately. The proposed threshold can be expressed as:
\begin{equation}
 \begin{split}
    & TH = \alpha + \beta E\CLB{\Tilde{x}_r}^3\CRB \\
    x_b & =
    \begin{cases} 
     +1,  \;\; if \;\; \Tilde{x}_r \geq TH \\ 
     -1,  \;\; if \;\; \Tilde{x}_r < TH
    \end{cases}.
\end{split}
    \label{eq:INSTA-Th}
 \end{equation}
We name this threshold as \textbf{INSTA}nce-aware \textbf{Th}reshold (\textbf{INSTA-Th}).
The INSTA-Th module is shown in \cref{fig:structure}a.

%------------------------------------------------------------------------
\subsection{Squeeze-and-Excitation Module}
\begin{figure}[t]
     %\centering
     \begin{center}
     \includegraphics[width=0.80\linewidth]{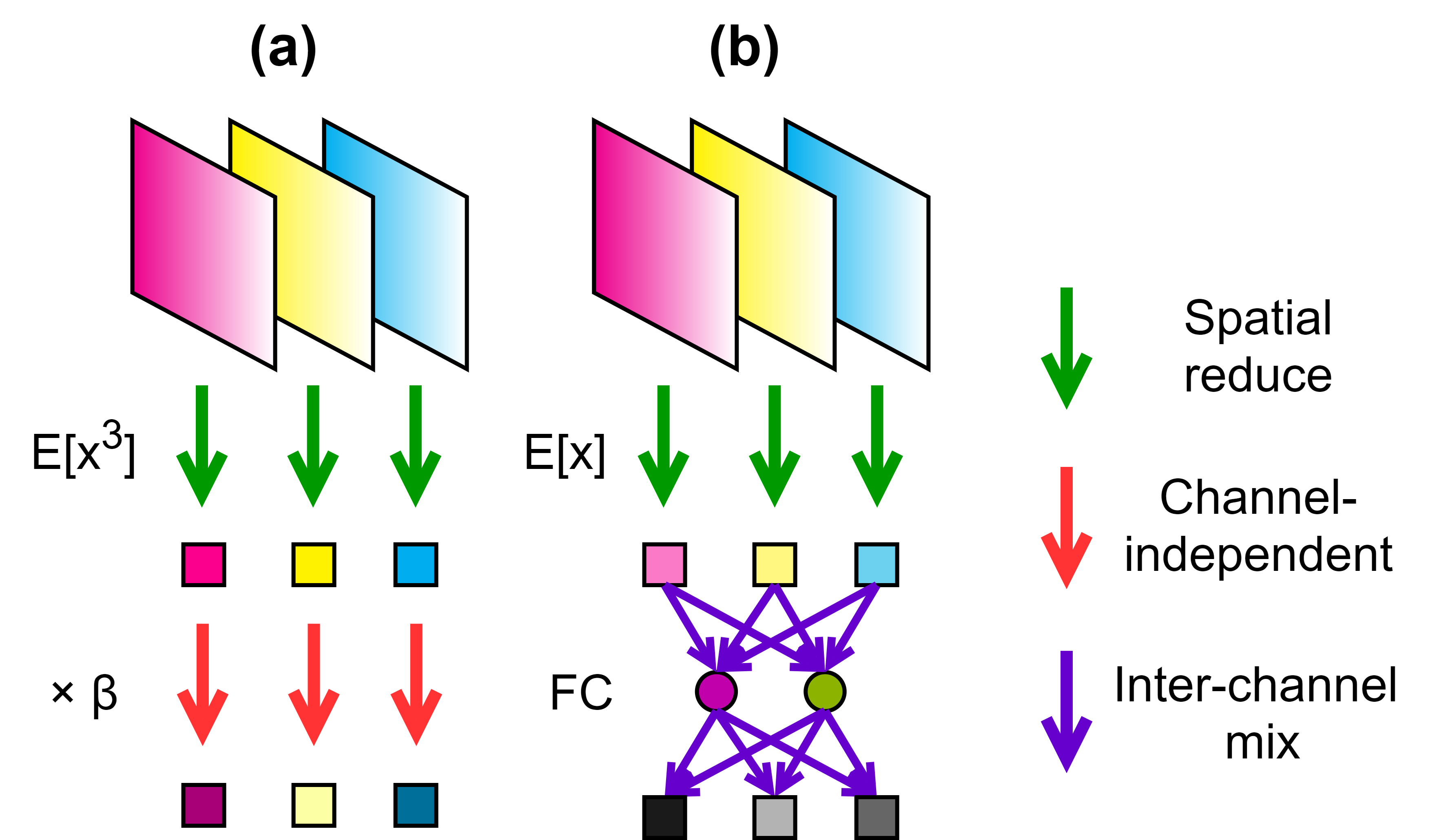}
     \end{center}
     \caption{Comparison of proposed INSTA-Th and SE module. (a) $\beta E\CLB\Tilde{x}_r^3\CRB$ component of INSTA-Th.  (b) SE module. Non-linear functions are skipped for simplicity.}
     \vspace{-2mm}
     \label{fig:compare_se}
 \end{figure}

Some of recent BNN works~\cite{martinez2020training} used the Squeeze-and-Excitation (SE) module~\cite{hu2018squeeze} to recalibrate the channel-wise magnitude based on a kind of instance-wise information.
Although the SE module and INSTA-Th may look similar in that both of them use instance-wise information, there is a notable difference.
As shown in \cref{fig:compare_se}, the SE module uses two fully-connected layers to assess the relative importance among different channels.
On the other hand, the proposed INSTA-Th computes $E\CLB X^3\CRB$ value channel-wise and then propagates the information to each channel independently.
Note that the INSTA-Th focuses on the difference between batch-wise and instance-wise statistical information rather than the difference between channels. 

Since the proposed INSTA-Th module and the SE module have different roles, 
the SE module can be combined with our INSTA modules for additional accuracy improvement.
In \cref{eq:INSTA-Th}, we modify the parameter $\alpha$ to be the output of the SE-like module instead of the channel-wise learnable parameter using back-propagation, as shown in \cref{fig:structure}b.
We replace the Sigmoid function in the original SE module with tanh function. To be more specific, we use the empirically found $3*$tanh($x/3$) to control the output range (Refer to Sec. F of the supplementary material for more detail).
In such a way, we intend $\alpha$ to focus on inter-channel relationship and $\beta E\CLB X^3\CRB$ to focus on instance-wise information.
Note that $\alpha$ is also an instance-aware variable in this case while it was a fixed value in previous works as in \cref{eq:rsign}.
We name the new threshold \textbf{INSTA-Th}$^{\bm{+}}$.
While INSTA-Th$^+$ requires a larger number of parameters than INSTA-Th, it can be clearly seen that using both INSTA-Th and SE-like modules improves the accuracy by a large margin (see \cref{section:imagenet_result}).
Therefore, we leave it as an option to select either INSTA-Th or INSTA-Th$^+$ depending on the requirement for the accuracy and parameter size.

%------------------------------------------------------------------------
\subsection{Instance-aware PReLU}
In recent works, additional activation functions such as ReLU and PReLU were used in BNNs for intermediate real-valued activations.
ReActNet used additional learnable parameters for these PReLU layers (RPReLU) as well as the Sign layers (RSign) as shown in \cref{fig:structure_ori_and_mean}a.
In addition to the INSTA-Th, we also propose replacing the learnable parameter (learnable shifts for x-axis) for RPReLU layers in ReActNet with instance-aware alternatives. We left the y-axis shift of RPReLU as it is.
\cref{fig:structure}a shows the proposed \textbf{INSTA}nce-aware \textbf{PReLU} (\textbf{INSTA-PReLU})  which replaces the PReLU.
In the INSTA-PReLU, we constrain the output range of $\beta E\CLB \Tilde{x}^3\CRB$. 
In case of the INSTA-Th, the output range was not important since the end layer of INSTA-Th was the Sign function.
However, as the PReLU layer does not constrain its output, extremely large values from the cubic operation may cause an unrecoverable problem in the network.
Therefore, we used the additional tanh function to constrain the output of the $\beta E\CLB \Tilde{x}^3\CRB$ term.
The computation process of the modified PReLU layer is as follows:
\begin{equation}
 \begin{split}
    \Tilde{x} = \frac{x - \hat{\mu}}{\hat{\sigma}} \qquad
    TH_{PR} = \alpha + tanh(\beta E\CLB \Tilde{x}^3\CRB) \\
    y = 
    \begin{cases} 
     \Tilde{x} - TH_{PR},  \qquad if \: \Tilde{x} \geq TH_{PR} \\ 
     \gamma(\Tilde{x} - TH_{PR}),  \:\:\: if \: \Tilde{x} < TH_{PR}
    \end{cases}.
\end{split}
    \label{eq:STAR-PR}
 \end{equation}
Here, $\gamma$ is a learnable slope parameter of the PReLU layer.
We can further control the output range of $\beta E\CLB \Tilde{x}^3\CRB$ term by replacing the tanh($x$) function with the $c*$tanh($x/c$).
We empirically found that $c$=3 worked well, so we chose to use $3*$tanh($x/3$) for all other experiments.
The effect of INSTA-PReLU will be discussed in \cref{section:component}.
Meanwhile, the parameter $\alpha$ in INSTA-PReLU (\cref{eq:STAR-PR}) can also be replaced by the SE-like module similar to the case in the previous section.
We call the module as \textbf{INSTA-PReLU}$^{\bm{+}}$.
In the remainder of the paper, we refer to the proposed BNN model as \textbf{INSTA}nce-aware \textbf{BNN} (\textbf{INSTA-BNN}), which employs both INSTA-Th and INSTA-PReLU modules.

%------------------------------------------------------------------------
\newcommand{\PLH}{{\mkern-2mu\times\mkern-2mu}}
\section{Practical guidelines for INSTA-BNN}
Although INSTA-Th and INSTA-PReLU modules show noticeable performance improvement (\cref{section:imagenet_result}), the cubic operations inside the modules cause non-negligible computational overhead. In this section, we introduce some techniques that we used to reduce the latency overhead of the proposed INSTA-BNN model while maintaining its benefit.

\begin{table}[t]
   \begin{center}
   \small
   \setlength\tabcolsep{3pt}
   \begin{tabular}{c||cccc||c}
     \hline
        & \multicolumn{4}{c||}{Layers using INSTA modules} & \multirow{3}{*}{\shortstack{Top-1 \\ Acc.(\%)}} \\
     \cline{1-5}
     layer name & conv2\_x & conv3\_x & conv4\_x & conv5\_x &  \\
     \cline{1-5}
     width & 64 & 128 & 256 & 512 & \\
     \hline
    original & \multirow{2}{*}{\checkmark} & \multirow{2}{*}{\checkmark} & \multirow{2}{*}{\checkmark} & \multirow{2}{*}{\checkmark} & \multirow{2}{*}{62.1} \\
    INSTA-BNN & & & & & \\
    \hline
    \multirow{8}{*}{\shortstack{INSTA-BNN \\ variants}} &  & \checkmark & \checkmark & \checkmark & 62.1 \\
    & \checkmark &  & \checkmark & \checkmark & 61.8 \\
    & \checkmark & \checkmark &  & \checkmark & 61.7 \\
    & \checkmark & \checkmark & \checkmark &  & 60.9 \\
    \cline{2-6}
    &  &  &  &  \checkmark & 61.1 \\
    &  &  &  \checkmark &  & 60.6 \\
    &  &  \checkmark &  &  & 60.6 \\
    &  \checkmark &  &  &  & 60.2 \\
    \hline
    Baseline & & & & & 60.0 \\
     \hline
   \end{tabular}
   \end{center}
   \caption{Variants of INSTA-BNN that selectively use INSTA modules and their ImageNet validation accuracy comparison.}
   \label{tab:skip}
\end{table}

\subsection{Selective use of the INSTA module}
In most convolutional neural networks (CNNs), the feature size decreases and the number of channels (width) increases in deeper layers. Meanwhile, the cubic operation within the INSTA module is an element-wise multiplication and its cost is proportional to the layer's feature size: $H \PLH W \PLH C$. 
As a result, the shallower layers have a larger cost for cubic operations, and the cost becomes smaller in deeper layers. Therefore, in terms of the amount of computation, it is more desirable to eliminate the cubic operation in earlier layers as far as accuracy is maintained.
\Cref{tab:skip} shows the accuracy comparison results for ResNet-18 based INSTA-BNN models in which the INSTA modules are applied to selective layers.
Interestingly, the use of INSTA modules at the deeper part of the model mainly accounts for the accuracy improvement.
When we apply INSTA modules to each block one by one, the results show a clear correlation and lead to a similar conclusion. 
Therefore, we can improve the accuracy of the model with relatively small computational overhead by skipping the INSTA modules for earlier layers.

\subsection{Reuse of activation statistics}

\begin{figure}[t]
    \centering
    \includegraphics[width=\linewidth]{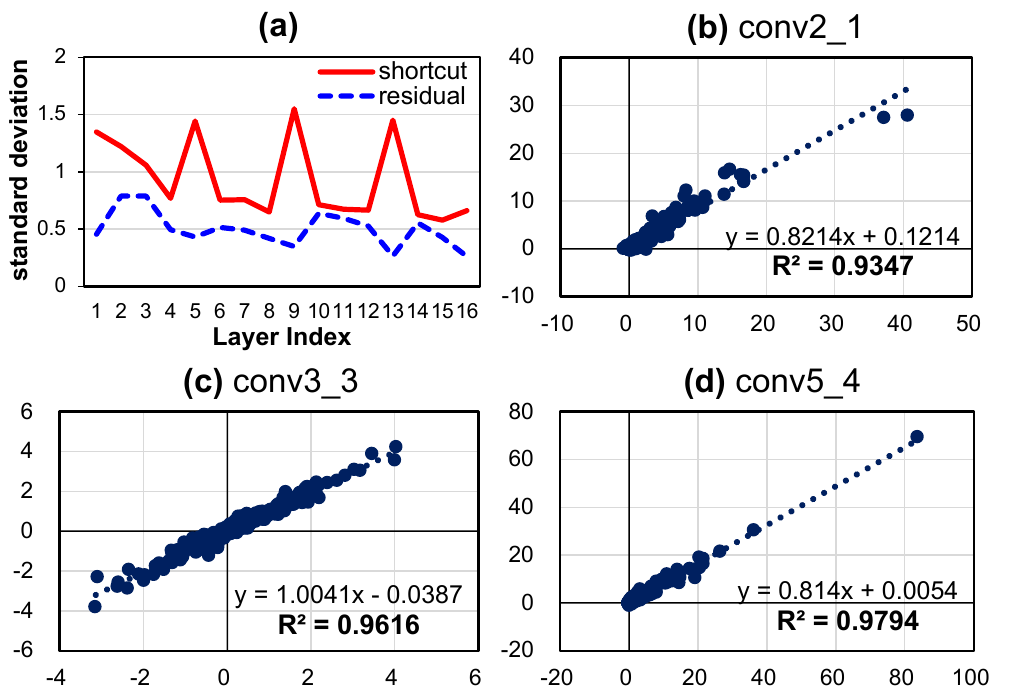}
    \caption{
    (a) The standard deviation of the values from the shortcut and residual right before they are added. (b)-(d) Correlation between $E\CLB \Tilde{x}^3\CRB$ values from INSTA-Th and subsequent INSTA-PReLU of 256 training set instances. x axis: $E\CLB \Tilde{x}^3\CRB$ values of INSTA-Th. y axis: $E\CLB \Tilde{x}^3\CRB$ values of subsequent INSTA-PReLU. Charts show the correlation at (b) the front, (c) the middle, and (d) the rear of the ResNet-18 based INSTA-BNN.
    }
    \label{fig:reuse}
    \vspace{-2mm}
\end{figure}

Since Bi-real net~\cite{liu2018bi}, most of BNNs have been adopting shortcuts for convolution blocks. 
In the proposed INSTA-BNN, the feature information used for INSTA-Th is directly transferred to INSTA-PReLU via an identity shortcut path while the binary convolution output is fed to INSTA-PReLU via the residual path at the same time (\cref{fig:structure}a). 
We compared the strength of the values from residual path with the ones from shortcut path by calculating their standard deviations (\cref{fig:reuse}a).
We can observe that the response strength of the shortcut path is larger than that of the residual path, which is consistent with the argument of the  ResNet~\cite{he2016deep} paper that the residual functions might have generally smaller values than the non-residual functions.

Based on the observation, we can expect that statistics ($E\CLB \Tilde{x}^3\CRB$) calculated from INSTA-PReLU module are similar to the statistics from the preceding INSTA-Th module, although they are not identical. We check their correlations for trained models. \cref{fig:reuse}b-d shows that calculated $E\CLB \Tilde{x}^3\CRB$ values inside INSTA-PReLU are highly correlated with previous $E\CLB \Tilde{x}^3\CRB$ of INSTA-Th module, whereas non-linear relations are observed between INSTA-PReLU and subsequent INSTA-Th because of the intervening PReLU (Sec. E in the supplementary material). Thus, we reuse the $E\CLB \Tilde{x}^3\CRB$ obtained from INSTA-Th for the subsequent INSTA-PReLU module to reduce the computational burden. 
In order to adjust the difference caused by the residual, we apply the channel-wise affine transform to the reused $E\CLB \Tilde{x}^3\CRB$ value.
When the shortcut contains 1x1 convolution, we calculate the feature statistics for INSTA-PReLU separately.
The experimental results show that applying the two schemes, the selective use of INSTA module and the reuse of feature statistics, do not degrade the accuracy while reducing the latency significantly. Detailed results are shown in \cref{section:imagenet_result} and \cref{section:latency}.
Please refer to the supplementary material Sec. D for the detailed model structure, which employs the two schemes.

%------------------------------------------------------------------------
\section{Experiments}
We applied the proposed schemes to state-of-the-art BNN models and evaluated the accuracy improvement on ImageNet (ILSVRC12) dataset~\cite{deng2009imagenet}.
%------------------------------------------------------------------------
\subsection{Experimental Setup}
\label{section:experi_setup}

\begin{table*}[t]
  \begin{center}
  \small
  \setlength\tabcolsep{11pt}
  \resizebox{\textwidth}{!}{
  \begin{tabular}{c||c||c|c|c|c||c|c}
    \hline
    \multirow{2}{*}{Method} & Bit-width & BOPs & FLOPs & OPs & Params & Top-1 & Top-5   \\
     & (W/A) & ($\times10^9$) & ($\times10^8$) & ($\times10^8$) & (Mbit) & Acc. (\%) & Acc. (\%)  \\
    \hline
    ResNet18 FP~\cite{pytorch2021pytorchzoo}   & 32/32 & 0 & 18.2 & 18.2 & 374 & 69.8 & 89.1 \\
    BNN~\cite{hubara2016binarized}         & 1/1 & 1.70 & 1.20 & 1.47 & 28.1 & 42.2 & 67.1 \\
    XNOR-Net~\cite{rastegari2016xnor}    & 1/1 & 1.68 & 1.40 & 1.66 & 33.4 & 51.2 & 73.2 \\
    Bi-real Net 18~\cite{liu2018bi}    & 1/1 & 1.68 & 1.40 & 1.67 & 33.5 & 56.4 & 79.5 \\
    XNOR-Net++~\cite{bulat2019xnor}    & 1/1 & 1.68 & 1.42 & 1.68 & 33.5 & 57.1 & 79.9 \\
    IR-Net~\cite{qin2020forward}    & 1/1 & 1.68 & 1.40 & 1.67 & 33.5 & 58.1 & 80.0 \\
    Bi-real Net 34~\cite{liu2018bi}   & 1/1 & 3.53 & 1.42 & 1.97 & 43.8 & 62.2 & 83.9\\
    Real-to-Binary~\cite{martinez2020training}    & 1/1  & 1.68 & 1.40 & 1.66 & 42.6 & 65.4 & 86.2 \\
    ReActNet-ResNet18~\cite{liu2020reactnet} \textbf{\textdaggerdbl}   & 1/1 & 1.68
    & 1.40 & 1.67 & 34.0 & 65.5 & - \\
    ReCU~\cite{xu2021recu}  & 1/1 & 1.68
    & 1.40 & 1.67 & 34.0 & 66.4 & 86.5 \\
    AdaBin~\cite{tu2022adabin} & 1/1 & 1.68 & 1.43 & 1.70 & 34.0 & 66.4 & 86.5 \\
    \textbf{INSTA-BNN-ResNet18}   & 1/1 & 1.68 & 1.43 & 1.70 & 34.8 & \textbf{67.6} & \textbf{87.5} \\
    \textbf{INSTA-BNN$^{\bm{+}}$-ResNet18}   & 1/1  & 1.68 & 1.44 & 1.70 & 37.3 & \textbf{68.5} & \textbf{88.2} \\
    \hline
    ReActNet-A~\cite{liu2020reactnet} \textbf{\textdaggerdbl}  & 1/1 & 4.82 & 0.12 & 0.87 & 63.1 & 69.4 & - \\
    ReActNet-B~\cite{liu2020reactnet}   & 1/1 & 4.69 & 0.44 & 1.17 & 68.0 & 70.1 & - \\
    AdaBin~\cite{tu2022adabin} & 1/1 & 4.82 & 0.21 & 0.96 & 63.1 & 70.4 & - \\
    \textbf{INSTA-BNN}  & 1/1 & 4.82 & 0.20 & 0.95 & 65.6 & \textbf{71.7} & \textbf{90.3} \\
    DyBNN~\cite{zhang2022dynamic}   & 1/1 & 4.82 & 0.14 & 0.90 & 137.4 & 71.2 & 89.8 \\
    High-Capacity Expert~\cite{bulat2021high}  & 1/1 & 1.70$^*$ & 1.10$^*$ & 1.37 & 83.0 & 71.2 & 90.1 \\
    ReActNet-C~\cite{liu2020reactnet}   & 1/1 & 4.69 & 1.40 & 2.14 & 84.7 & 71.4 & - \\
    \textbf{INSTA-BNN$^{\bm{+}}$}  & 1/1 & 4.82 & 0.20 & 0.96 & 71.2 & \textbf{72.2} & \textbf{90.5} \\
    \hline
  \end{tabular}
  }
  \end{center}
  \vspace{-1mm}
  \caption{
  Comparison of the number of operations, parameter size, and ImageNet top-1 and top-5 validation accuracy of different BNN models.
  BOPs and FLOPs mean the number of binary operations and floating point operations, respectively. $^*$ indicates that those numbers are from their original paper~\cite{bulat2021high}. \textbf{\textdaggerdbl}~denotes the baseline model of the proposed INSTA-BNN(-ResNet18) models.
  }
  \vspace{-2mm}
  \label{tab:imagenet_top1}
\end{table*}

We used the PyTorch framework to implement state-of-the-art BNN models and the proposed INSTA-BNNs.
We chose the ReActNet~\cite{liu2020reactnet} as the baseline model.
We replaced the RSign layer in ReActNet with INSTA-Th (or INSTA-Th$^+$) module and the PReLU layer with INSTA-PReLU (or INSTA-PReLU$^+$) module. The reduction ratio of 16 is used for INSTA-Th$^+$ and INSTA-PReLU$^+$, if not specified.
We followed the two-stage training strategy in~\cite{martinez2020training}.
In the first stage, we trained a model with binary activation and full-precision weights from scratch.
In the second stage, we trained a BNN model by initializing it with pre-trained weights from the first stage.
We used Adam optimizer with a linear learning rate decay scheduler. In both stages, we trained the model for 256 epochs with batch size of 512 and initial learning rate of 0.001. We only applied random resized crop and random horizontal flip for the training data augmentation and used default STE~\cite{bengio2013estimating} rather than a piecewise polynomial function from~\cite{liu2018bi}.
We also used the distributional loss~\cite{liu2020reactnet} using ResNet-34 as a teacher model. Note that our proposed INSTA-BNN does not rely on any specific BNN training methods. 
To reduce the additional parameter cost for the proposed scheme, we applied the 8-bit quantization to the weights of the SE-like module. Please refer to the supplementary materials for more details on weight quantization and cost analysis.
%------------------------------------------------------------------------
\subsection{Comparison on ImageNet Classification}
\label{section:imagenet_result}

In this section, we compare the accuracy and cost of previous BNN models and the proposed INSTA-BNN models.
In \Cref{tab:imagenet_top1}, we grouped the models with similar model parameter sizes.
For the proposed models, INSTA-BNN represents a model with INSTA-Th and INSTA-PReLU modules, and INSTA-BNN$^+$ represents a model with INSTA-Th$^+$ and INSTA-PReLU$^+$ modules.

INSTA-BNN-ResNet18 / INSTA-BNN$^+$-ResNet18 use the ReActNet-ResNet18 model as a backbone network while INSTA-BNN / INSTA-BNN$^+$ use the ReActNet-A (based on MobileNetV1) model as a backbone network.
For experimental results, INSTA-BNN$^+$ (bottom row of \Cref{tab:imagenet_top1}) used the INSTA-Th$^+$ and INSTA-PReLU (without SE modules) to avoid large increase in the parameter size.

Compared to BNN models with less than 50 Mbit parameters, the proposed INSTA-BNNs with ResNet-18 backbone achieved much higher accuracy.
With a small overhead, INSTA-BNN with ResNet-18 backbone achieved 67.6\% top-1 accuracy which is already 2.1\% higher than ResNet-18 based ReActNet. 
In addition, with additional parameter cost, INSTA-BNN$^+$ with ResNet-18 backbone achieved even higher accuracy (68.5\% top-1 accuracy) which is 3.0\% higher than baseline, and only 1.3\% lower than Full-Precision ResNet-18 baseline.
Similar improvement was achieved with larger BNN models.
INSTA-BNN achieved 2.3\% higher accuracy than ReActNet-A with marginal overhead.
In case of INSTA-BNN$^+$, it achieved much higher top-1 accuracy than the state-of-the-art models, ReActNet-C~\cite{liu2020reactnet} and High-Capacity Expert~\cite{bulat2021high}, with much smaller parameter size and computing cost.

We trained the full-precision (fp) network with ReActNet structure using the official ReActNet implementation~\cite{reactnet2020} and got 73.1\% top-1 accuracy. Applying weight decaying not only to the convolution layers but also to the linear weights prevented the overfitting of fp model training, resulting in higher accuracy than originally reported. As a result, there is a reasonable gap between ReActNet fp model and INSTA-BNN$^{\bm{+}}$ accuracy.
\begin{figure}[t]
    \centering
    \includegraphics[width=\linewidth]{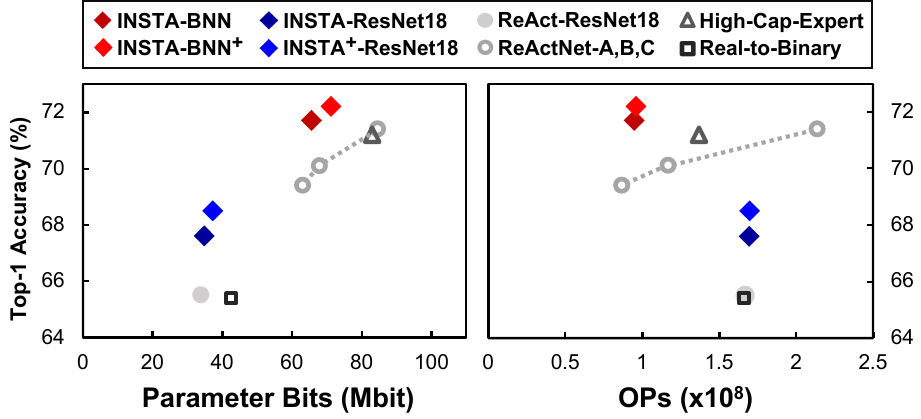}
    \caption{
    (a) Parameter size versus ImageNet top-1 accuracy. 
    (b) The number of operations (OPs) versus ImageNet top-1 accuracy. 
    }
    \label{fig:pareto_plot}
\end{figure}
The superiority of the proposed models is more clearly demonstrated in \cref{fig:pareto_plot}.
The proposed INSTA-BNN models are placed in the upper-left side of previous state-of-the-art models in both accuracy vs. parameter size graph and accuracy vs. operations graph.

%------------------------------------------------------------------------
\subsection{Inference latency evaluation}
\label{section:latency}
In the previous section, we showed that INSTA-BNNs produce higher accuracy with small operation overhead.
Similar to our analysis in previous section, previous works mostly reported the number of operations (OPs) to assess the computational overhead, but simple counting of OPs may not directly translate into real hardware latency characteristics.
So, we also evaluated the latency of BNNs (\Cref{tab:latency}) using the Huawei Noah's Ark Lab's Bolt~\cite{noah2020} library on Google pixel 3, which has 2.80 GHz Qualcomm Kryo 385 Gold and 1.77 GHz Silver (customized ARM Cortex-A75 and A55, respectively) CPUs inside.
We customized the bolt kernel to make cubic operations and element-wise addition more computationally efficient. As shown in \Cref{tab:latency}, INSTA-BNNs have marginal overhead compared to their baselines. We report the average latency of 10 inference loops in the table. 
Using a single thread / Cortex-A75 core, INSTA-BNN runs with just 2 ms (6\%) and 3 ms (4\%) latency overhead compared to ResNet-18 and MobileNetV1 baselines, respectively, while maintaining the higher accuracy.
This shows that INSTA-BNN can operate efficiently on real hardware while achieving higher accuracy.

\begin{table}[t]
   \begin{center}
   \small
   \setlength\tabcolsep{4pt}
   
   \begin{tabular}{c||c||c|c|c}
     \hline
        \multirow{3}{*}{Model} & Top-1 & .bolt & A75 & A55 \\
        & Acc. & file size & latency & latency  \\
        & (\%) & (MB) & (ms) & (ms) \\
     \hline
    Bi-real Net 18      & 56.4 & 2.68 & 30 & 83     \\
    Bi-real Net 34      & 62.2 & 3.91 & 52 & 135     \\
    Real-to-Binary      & 65.4 & 3.34 & 33 & 93     \\
    ReActNet-RN18   & 65.5 & 2.74 & 32 & 91     \\
    \textbf{INSTA-BNN-RN18}  & 67.6 & 2.85 & 34 & 98      \\
    \textbf{INSTA-BNN$^{\bm{+}}$-RN18} & 68.5 & 3.55 & 35 & 102  \\
    \hline
    ReActNet-A & 69.4 & 5.55 & 70 & 178 \\
    \textbf{INSTA-BNN} & 71.7 & 5.79 & 73 & 192  \\
    \textbf{INSTA-BNN$^{\bm{+}}$} & 72.2 & 7.19 & 75 & 197 \\
     \hline
   \end{tabular}
   \end{center}
   \caption{Inference time comparison of BNNs on mobile device. Measured using a single thread. .bolt is a converted file that can be used for inference.
   RN18 = ResNet18.}
   \label{tab:latency}
\end{table}

%------------------------------------------------------------------------
\subsection{Ablation Study}
\label{section:ablation}

\subsubsection{Effect of individual component in INSTA-BNN}
\label{section:component}
\Cref{tab:comp_contribution} shows the effect of each component proposed in \cref{section:insta-bnn} on the ImageNet top-1 accuracy. 
Experimental setup for the ablation study is described in Sec. A.2 of the supplementary material.
First, using the instance-wise mean information on the threshold (\cref{eq:th_mean}) improves the top-1 accuracy by 0.6\% when compared to the baseline ReActNet model.
On top of that, when variance and skewness information are added on the threshold (INSTA-Th), another 0.5\% improvement is achieved.
Replacing the learnable parameter $\alpha$ in \cref{eq:INSTA-Th} with the SE-like module with reduction ratio ($r$) of 8 achieves 61.7\% top-1 accuracy that is 1.7\% higher than baseline result.
We also evaluate different combinations of INSTA-Th and INSTA-PReLU with and without the SE-like modules.
When both modules use the SE-like module (INSTA-Th$^+$ and INSTA-PReLU$^+$), we increase the $r$ to 16 to match the additional parameter cost.
Regardless of where the SE-like module is combined with INSTA modules, it improves the accuracy by a large margin. However, the cases in which only SE module is used in the proposed architecture ($2nd$ and $3rd$ row of \Cref{tab:comp_contribution}) show lower accuracy than the cases when INSTA modules are used.

\begin{table}[t]
  \begin{center}
  \small
  \setlength\tabcolsep{3pt}
  
  \begin{tabular}{lc}
    \hline
    Network                     & Top-1 Acc. (\%) \\
    \hline
    \hline
    Baseline \qquad \space (RSign \space\space + RPReLU)   & 60.0 \\
    \hline
    Baseline + SE (RSign$^+$ + RPReLU) \space\space\space (r=8)  & 60.8  \\
    Baseline + SE (RSign$^+$ + RPReLU$^+$) (r=16) & 61.1\\
    \hline
    Normalization and mean (\cref{eq:th_mean})     & 60.6 \\
    
    INSTA-Th                         & 61.1 \\
    INSTA-Th \textcolor{red}{$-$norm}    &  60.8     \\
    INSTA-Th$^+$ (r=8)                & 61.7 \\
    
    INSTA-Th \space\space  + \space INSTA-PReLU     & 62.1 \\
    INSTA-Th \space\space + \space (INSTA-PReLU \textcolor{red}{$-$norm}) & 61.5 \\
    INSTA-Th$^+$ + \space INSTA-PReLU \space\space (r=8)     & 63.0 \\
    INSTA-Th \space\space+ \space INSTA-PReLU$^+$ (r=8)    & 62.8 \\
    INSTA-Th$^+$ + \space INSTA-PReLU$^+$ (r=16)    & 62.9 \\
    \hline
  \end{tabular}
  \end{center}
  \caption{The effect of each component of INSTA-BNN on the ImageNet top-1 validation accuracy. 
  $^+$ and r indicate the usage of SE module and the reduction ratio, respectively.}
  \label{tab:comp_contribution}
\end{table}

\subsubsection{Effect of the normalization layer}
In \cref{section:instance}, we proposed using the normalization layer before calculating instance-wise mean information to capture the difference between instance-wise and batch-wise statistical data.
Although the proposed modules can still make instance-aware thresholds without the normalization layer, we observed that removing the normalization from either INSTA-Th or INSTA-PReLU modules limits the performance of the modules.
When normalization layers are not used (\textcolor{red}{$-$norm} indication in the \Cref{tab:comp_contribution}), the modules generate the threshold using the instance-wise statistical data themselves.
In contrast, when the normalization layers are used, the difference between the instance-wise and overall training dataset statistics are used to generate the threshold ($5th$ row in \Cref{tab:comp_contribution}), resulting in better accuracy than using pure instance-wise statistics ($6th$ row).

Ablation study for the effect of the threshold value ranges in INSTA modules, the effect of INSTA-Th for different block structure, discussion of reducing the inconsistent sign problem of binary convolution, and visualization results of t-SNE are discussed in detail in the supplementary materials. 

%------------------------------------------------------------------------
\section{Conclusion}

In this paper, we argue that the traditional BNNs with the input-agnostic threshold are sub-optimal. Instead, we demonstrate that the instance-wise statistics, including mean, variance, and skewness, must be considered to determine the better threshold values dynamically. 
Based on the idea, we propose the BNN with instance-aware threshold control (INSTA-Th) and demonstrate that the proposed BNN outperforms the previous BNNs by a large margin. 
We further improve the performance of the BNN with INSTA-Th by adding the instance-aware PReLU (INSTA-PReLU) and a variant of the Squeeze-and-Excitation module (INSTA-Th$^+$).
Experimental results show that our INSTA-BNN$^+$ achieves the top-1 accuracy up to 72.2\% on the ImageNet dataset.

\newpage

\section*{Acknowledgement}
This work was supported in part by Institute of Information \& communications Technology Planning \& Evaluation (IITP) grant funded by the Korea government (MSIT) (No. 2021-0-00105, Development of model compression framework for scalable on-device AI computing on Edge applications, and No. 2021-0-02068, Artificial Intelligence Innovation Hub).

%------------------------------------------------------------------------

{\small
\bibliographystyle{ieee_fullname}

}

\newpage

\section*{A\quad Experimental setups}
\subsection*{1\quad Experimental setup for CIFAR-10 dataset}
Extensive experiments for \mbox{Fig. 3} in the main paper were conducted using ResNet-20~\cite{he2016deep} with CIFAR-10 dataset. More precisely, we used ResNet-20 with additional skip connections from~\cite{liu2018bi} and attached PReLU after the element-wise addition of residual path and identity shortcut. Therefore, the convolution block structure is similar to the block used in ReActNet~\cite{liu2020reactnet} (Fig. 2a of the main paper).
We used AdamW optimizer~\cite{loshchilov2017decoupled} with a weight decay value of 1e-4, and the networks were trained up to 400 epochs. The initial learning rate was 0.003, and the cosine annealing schedule was used. The batch size was set to 256. We used the weight scaling factor from~\cite{rastegari2016xnor}.

\subsection*{2\quad Experimental setup for the ablation study}
We used a simplified training process for the ablation studies, in which 
we trained a BNN model directly from scratch for 90 epochs without help of the pre-trained or teacher models. 
Adam optimizer~\cite{kingma2014adam} was used with an initial learning rate of 0.001, which is multiplied by 0.1 at epochs 40, 60, 80.
Weight decay was not used, and learning rate warmup was used for the first 5 epochs. 
Precision other than 1-bit and 32-bit was not used for the ablation study.
Experiments that produced the results in Table 1 of the main paper and the rest of the supplementary material experiments (Sec. F, G, H) also followed this setup.

\section*{B\quad Details and training methods for the quantization of SE-like modules}
As mentioned in Sec. 5.1 of the main paper, we quantized the weights of the SE-like modules.
We followed the fine-tuning scheme of learned step size quantization (LSQ)~\cite{esser2019learned}. We used channel-wise learnable step size for 8-bit weight quantization and symmetric quantization ($Q_N = 2^{b-1}$ and $Q_P = 2^{b-1} - 1$, encoding with $b$ bits) was applied. 

After the two-stage training, we obtain the conventional BNN with binary convolution blocks and all other real-valued components.  
When the SE module is used, additional 10 epochs of fine-tuning were performed with 8-bit SE weights. The initial learning rate was 1e-4 and we used a linear learning rate decay scheduler. We did not use the weight decay for fine-tuning.
This 8-bit SE weight quantization did not cause accuracy degradation in our experiments.

\begin{figure*}[t]
    \centering
    \includegraphics[width=\linewidth]{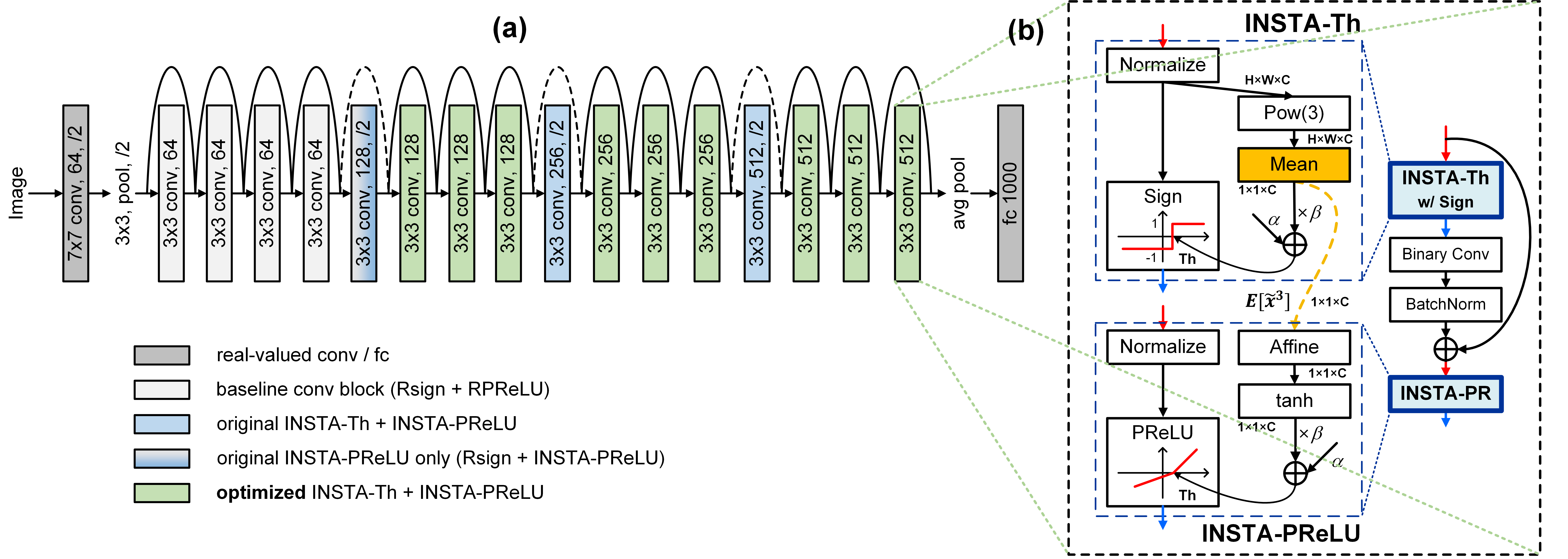}
    \caption{(a) Block diagram of latency-optimized INSTA-BNN-ResNet18. (b) Binary convolution block that contains the optimized INSTA-Th and INSTA-PReLU modules. Orange block indicates the calculated $E\CLB \Tilde{x}^3\CRB$ value in the INSTA-Th module.  We can see that $E\CLB \Tilde{x}^3\CRB$ value is reused for the INSTA-PReLU (indicated by the yellow dotted line).}
    \label{fig:optimized}
\end{figure*}

\section*{C\quad Detailed rules for the computational cost analysis}

In INSTA-BNN, the additional operations are mainly from normalizing and cubic operations.
We calculated total number of opertations (OPs) as OPs = FLOPs + (BOPs / 64), following ~\cite{liu2020reactnet,liu2018bi}.
In case of parameters, binary weights are 1-bit, weights of SE-like modules are 8-bit, and other real-valued parameters and weights are considered as 32-bit.

We manually calculated the computation cost and parameter size of the previous methods and our proposed INSTA-BNN. In this section, we introduce additional detailed rules that we used to calculate the computation cost of the network. 
 First, we only counted the number of floating point (FP) multiplication for the FLOPs calculation following the method used in~\cite{liu2020reactnet,liu2018bi}. That is, we count one floating point multiply-accumulate (MAC) as one FLOP, so our FLOPs calculation includes a pair of multiplication and addition.
 The same rule applies to the calculation of BOPs.

 For the Batch Normalization layer, we assumed that the layer has $2\PLH C$ sets of parameters because four BN parameters can be merged into the scale and shift terms offline. FP multiplication cost of BN layer is $H \PLH W \PLH C$. However, when the Sign function directly comes after BN layer, FP multiplication cost is removed and we only need $1\PLH C$ set of parameters, following~\cite{kim2019memory}. In case of using weight scaling factor~\cite{rastegari2016xnor}, the element-wise FP multiplication cost of $H_{out} \PLH W_{out} \PLH C_{out}$ is required, where the subscript $out$ means the output of the convolution. However, when a method has conv-BN layer ordering, instead of doing element-wise multiplication for each scaling factor and BN layer, we can first multiply analytically calculated scaling factor and the merged BN scale parameter. In this way, we can also remove one set of $H \PLH W \PLH C$ element-wise multiplications.
 We ignored the operation cost for PReLU layer because the cost may vary depending on the implementation, and its effect is very small compared to the real-valued convolution cost. 
 
 Finally, we assumed that all the ResNet-based methods use the shortcut option B reported in \cite{he2016deep}. In addition, we assumed that average pooling, real-valued 1x1 convolution, and BN layer are used for the downsampling path, except the BNN~\cite{hubara2016binarized} case, which uses binary 1x1 convolution.

 Our proposed INSTA-Th needs $4\PLH C$ set of parameters ($\hat{\mu}$, $\hat{\sigma}$, $\alpha$, $\beta$). For INSTA-PReLU, total five parameters per channel (including the learned slope of PReLU) are required. Both modules need $H \PLH W \PLH C$ floating point operations for normalization and $2 \PLH H \PLH W \PLH C$ floating point operations for cubic operation. 
 For INSTA-Th, additional $2\PLH C$ operations are needed for spatial averaging and multiplying channel-wise parameters.
 In addition, for INSTA-PReLU, $3\PLH C$ operations are needed because we additionally use 3*tanh($x/3$) for INSTA-PReLU. 
 Note that the computational cost of these operations is proportional to the channel size, which is negligible compared to the computational cost of the element-wise operations. 
 
 For the optimized INSTA-BNN structure, the feature statistics reuse (Sec. 4.2 of the main paper) eliminates a cubic operation for the corresponding INSTA-PReLU module. It also removes spatial averaging thereby reducing the number of required extra operations from $3\PLH C$ to $2\PLH C$. Instead, two channel-wise parameters are used for the affine transform of the reused statistics (\cref{fig:optimized}b).
 
 INSTA-Th$^+$ and INSTA-PReLU$^+$ further require $2C^2/r$ 8-bit parameters for fully-connected layers of each SE-like module ($r$: reduction ratio) instead of one channel-wise parameter ($\alpha$).
 We also included the additional computational cost caused by the SE weight quantization in total OPs calculation. Due to the channel-wise step size of 8-bit weight quantization, we need $C+C/r$ real-valued parameters and multiplications for each SE-like module.

\begin{figure*}[t]
    \centering
    \includegraphics[width=\linewidth]{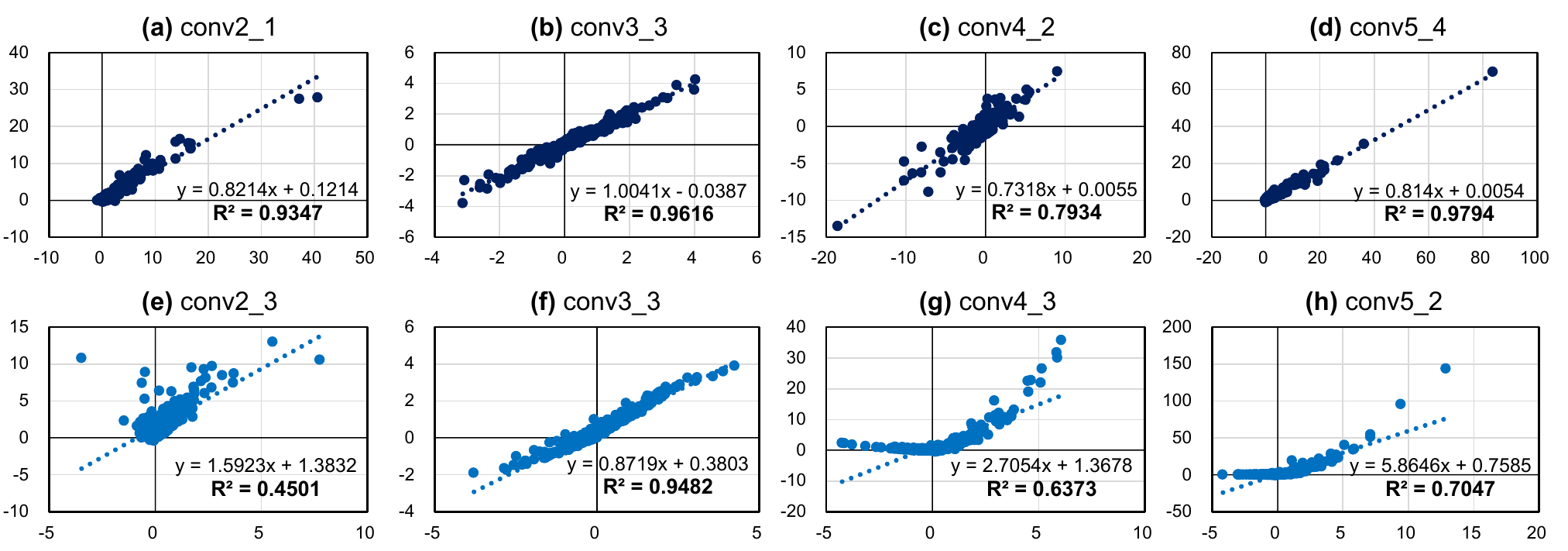}
    \caption{(a)-(d) Correlation between $E\CLB \Tilde{x}^3\CRB$ values from INSTA-Th and subsequent INSTA-PReLU.
    (e)-(h) Correlation between $E\CLB \Tilde{x}^3\CRB$ values from INSTA-PReLU and next conv block's INSTA-Th.}
    \label{fig:reuse_all}
\end{figure*}

\section*{D\quad Detailed model structure of latency-optimized INSTA-BNN}
For latency-optimized INSTA-BNN models, we used the INSTA modules from the point where the feature size is halved for the third time in the model structure ($H = 28$ when input image size is 224).  \cref{fig:optimized} shows the optimized model structure of INSTA-BNN-ResNet18. In \cref{fig:optimized}a, although the RPReLU (INSTA-PReLU) of each conv block comes after a shortcut is added, the diagram is presented differently for simplicity. These latency reduction schemes are similarly applied to the MobileNet-based INSTA-BNN.

\section*{E\quad Analysis of feature statistics reuse in INSTA module}
\cref{fig:reuse_all} shows the relationships between statistics ($E\CLB \Tilde{x}^3\CRB$) of INSTA-Th and the subsequent INSTA-PReLU (\cref{fig:reuse_all}a-d) introduced in the main text, as well as the correlation between $E\CLB \Tilde{x}^3\CRB$ value of INSTA-PReLU and next conv block's INSTA-Th (\cref{fig:reuse_all}e-h). Each subgraph shows the correlation between a pair of internal INSTA modules for each ResNet block.
We compared the $E\CLB {x}^3\CRB$ value calculated immediately after normalization for 256 training set images. Figures are drawn using channel 0 data in this example.
We can observe that statistics ($E\CLB \Tilde{x}^3\CRB$) calculated from INSTA-Th and those from the subsequent INSTA-PReLU have high correlation (\cref{fig:reuse_all}a-d) so that we can reuse the statistics obtained from INSTA-Th for INSTA-PReLU.
In contrast, \cref{fig:reuse_all}g, h show some problematic cases, in which significant non-linearity is observed due to the intervening PReLU. Note that there is a large difference between the x-axis and y-axis value scales. 
For this reason, we do not reuse the statistics obtained from INSTA-PReLU for INSTA-Th of the next conv block.

\section*{F\quad Effect of the threshold value range in INSTA-PReLU/INSTA-TH$^+$}
\label{section:output_range}
 \begin{table}[t]
   \begin{center}
   \setlength\tabcolsep{3pt}
   
   \begin{tabular}{c||c||c|c}
     \hline
         & Range & INSTA-PReLU & SE (INSTA-Th$^+$) \\
     \hline
     $sigmoid(x)$ & 0 $\sim$ 1 & 61.8      &  61.7     \\
     $tanh(x)$    & -1 $\sim$ 1    & 61.9      &  61.4     \\
     $3tanh(x/3)$ & -3 $\sim$ 3 & \textbf{62.1}      & \textbf{61.7}      \\
     \hline
   \end{tabular}
   \end{center}
   \caption{ImageNet top-1 valid accuracy (\%) according to the non-linear component of the each INSTA-PReLU and SE of INSTA-Th$^+$ module. Range means possible output range of each function.}
   \vspace{-1mm}
   \label{tab:last_module}
 \end{table}

 We use the bounded non-linear functions in the INSTA-PReLU, INSTA-Th$^+$, and INSTA-PReLU$^+$ modules to map the threshold values in the constrained range (Fig. 4a, b in the main paper). 
 We evaluated three options; sigmoid($x$), tanh($x$), and 3*tanh($x/3$). Sigmoid is a widely used function and was used in the original SE work. However, the output of the sigmoid is always greater than zero, and hence it can only move the threshold to one direction from zero. On the other hand, tanh is bidirectional. 
 We observed that the max value of $\alpha$ in Eq. (8) for the trained networks was 2.45, so that we included 3*tanh($x/3$) in the options.
 From the \Cref{tab:last_module}, 3*tanh($x/3$) was found to be the best non-linear function for the INSTA-PReLU, and it showed similar accuracy to that of sigmoid for SE module. Based on the results of \Cref{tab:last_module} and the observed $\alpha$ ranges, we used 3*tanh($x/3$) for the experiments in this paper.

\section*{G\quad INSTA-Th for different block structure}

\begin{table}
  \begin{center}
  
  \begin{tabular}{l c }
    \hline
        & Top-1 Acc. (\%) \\
    \hline
    BN - sign - conv - PReLU   & 59.5\\
    BN (w/o $\gamma$, $\beta$) - sign - conv - PReLU      & 59.0\\
    INSTA-Th \space - \space conv - PReLU  & 60.5\\
    \hline
  \end{tabular}
  \end{center}
  \caption{
  ImageNet top-1 accuracy comparison between the BN-sign-conv layer ordered network with and without INSTA-Th module. When the INSTA-Th module is used, its normalization layer replaces a BN layer of the network.}
  \label{tab:bn_sign}
\end{table}

In recent studies~\cite{bulat2021high,martinez2020training}, a different convolution block layer order from the one used in the Bi-real-net~\cite{liu2018bi} has been used. Its layer order is BatchNorm (BN)-Sign-conv-PReLU-$\oplus$, where $\oplus$ means element-wise addition of residual path and identity shortcut starting just before BN. To verify the versatility of the proposed method, we also tested INSTA-Th to this block structure.
One thing to note is that the normalization layer inside the newly added INSTA-Th comes immediately after the BN of the original structure (BN - Normalization - Sign with modified $TH$). In this case, the scale and shift effect of the original BN layer may be inhibited, and the meaning of the added normalization layer is also faded.
To handle this, we tried to merge BN and the newly added normalization layer (replacing the original BN with the proposed INSTA-Th), and its result is in \Cref{tab:bn_sign}. 
Removing the scale and shift from the original BN brought a slight decrease in accuracy (row 1 and 2), but merging BN and normalization layer of INSTA-Th shows the improved threshold controlling ability (row 1 and 3). 
This suggests that the proposed INSTA-Th plays a similar role to the shift of BN, but it works instance-wise and improves the performance further. 
This result supports the effectiveness of the proposed INSTA-Th in various block structures.

\begin{figure}[t]
    \centering
    \includegraphics[width=0.95\linewidth]{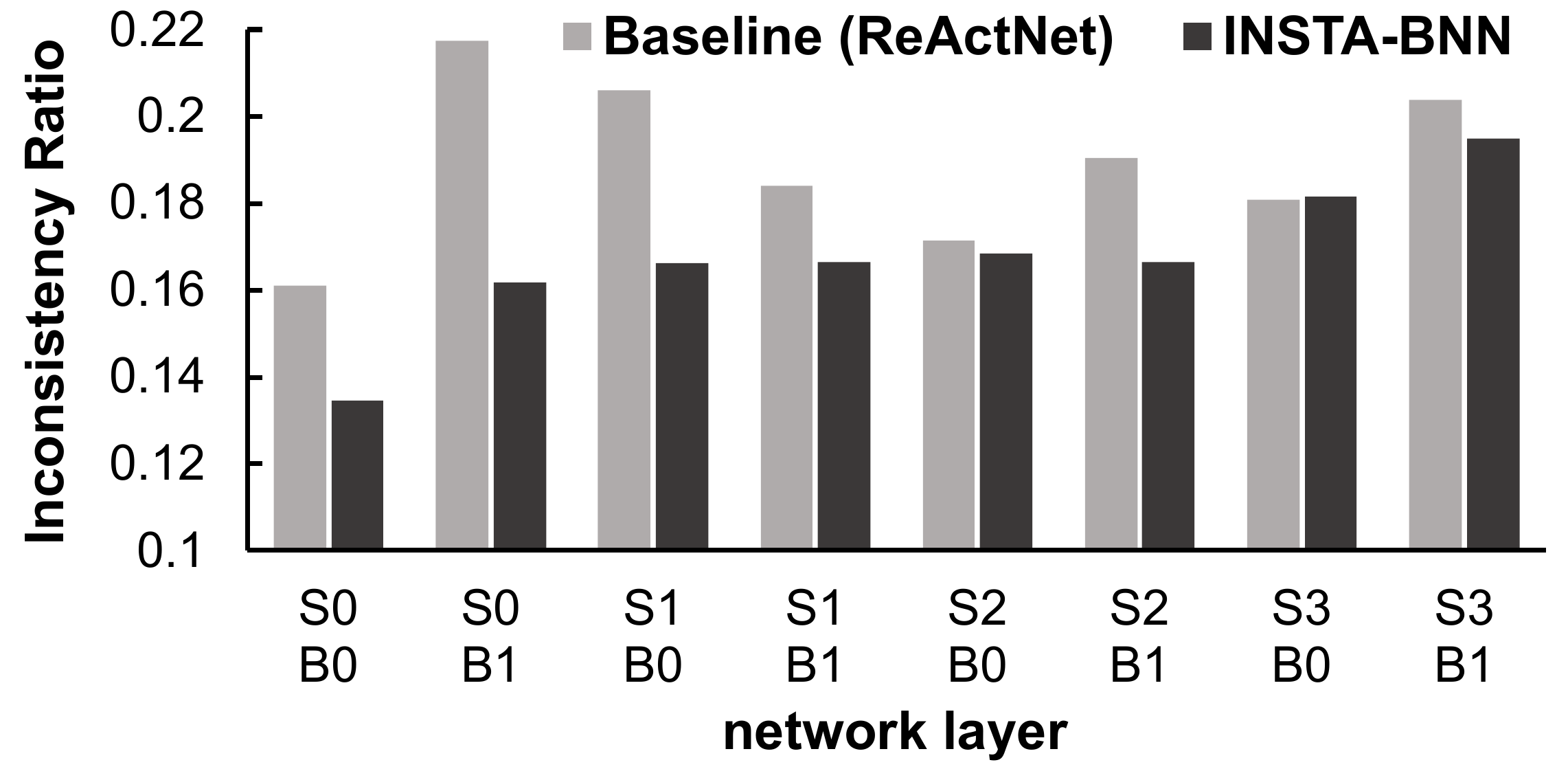}
    \caption{
    Comparison of the inconsistent sign ratio between the baseline and the proposed INSTA-BNN. S and B represent the stage and the block of ResNet structure. Note that ResNet-18 consists of four stages with two blocks each, and therefore S0 B0 contains conv2 and conv3 layers.}
    \label{fig:ci-bcnn}
\end{figure}

\section*{H\quad Discussion: Reducing the inconsistent sign problem of binary convolution}
CI-BCNN~\cite{wang2019learning} discussed that one of the key reasons for the quantization errors in BNN is the sign mismatch between the binary convolution results and the counterpart full-precision results $(sign(W_r \otimes A_r) \neq sign(W_b \oplus A_b))$. 
Then they tried to solve the inconsistency by imposing the channel-wise priors on the feature maps. 
Meanwhile, tuning the threshold is a direct way of changing the sign of binary activation, which consequently affects the sign of the binary convolution output. 
Hence, we checked the inconsistent ratio (ratio of pixels with inconsistent signs) for our baseline ReActNet and the proposed INSTA-BNN, and observed that the ratio was consistently lower in the INSTA-BNN (\cref{fig:ci-bcnn}).
The results demonstrate that our INSTA-BNN has the ability to reduce the quantization error by controlling the binary threshold without using channel-wise priors.

\section*{I\quad Visualization results of t-SNE}

\begin{figure}[t]
    \centering
    \includegraphics[width=0.95\linewidth]{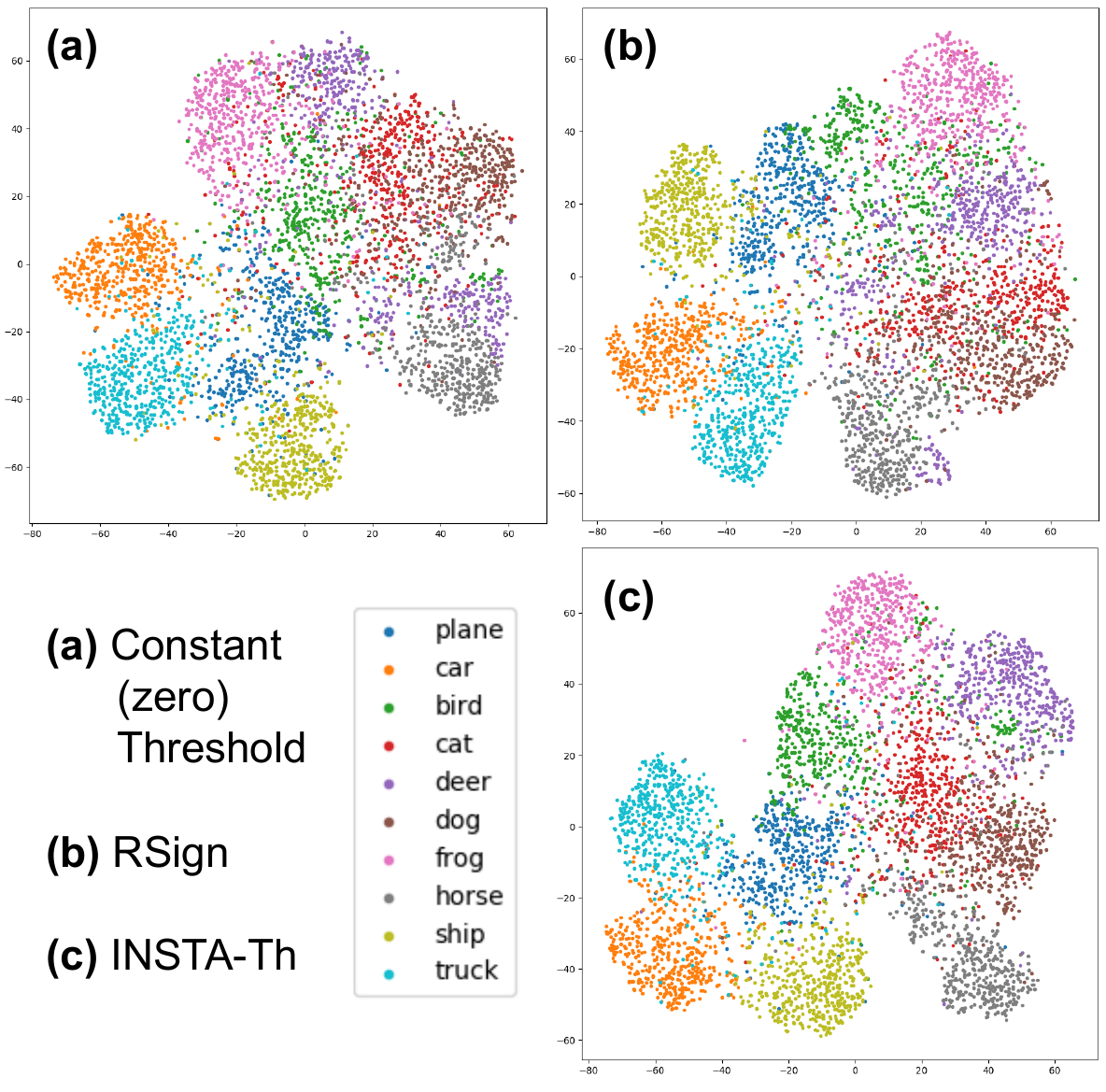}
    \caption{t-SNE visualization of features extracted before the classifier  (fully-connected layer) of each model. Half of the CIFAR-10 test set was used (500 images per class). Best viewed in color.}
    \label{fig:t-SNE}

\end{figure}

We compared three models based on ResNet-20 having different threshold determination methods: Sign function with a constant threshold of zero, RSign~\cite{liu2020reactnet} function, and INSTA-Th.
\cref{fig:t-SNE} shows t-SNE visualization of features extracted before the classifier of each trained model. 
For the INSTA-Th case (\cref{fig:t-SNE}c), cat (red) and dog (brown) classes, which are relatively difficult to distinguish in other models, are better distinguished.
Also, deer (purple) and bird (green) are relatively well clustered in the model with INSTA-Th. 

\section*{J\quad Comparison with DyBNN}
DyBNN~\cite{zhang2022dynamic} and INSTA-BNN have an apparent difference in that DyBNN relies solely on the channel-wise mean, while we employ higher-order statistics as well. 
Fig. 1 of the main paper shows the motivation for this: the higher-order statistics clearly describe the characteristics of the pre-activation distribution. Based on this, 
INSTA-BNN determines a more appropriate threshold for each instance.

Please note that basic INSTA-BNN outperforms DyBNN (71.7\% vs. 71.2\%) without extra parameters of Fully-Connected (FC) modules. Moreover, INSTA-BNN$^{\bm{+}}$ (72.2\%) further improves accuracy by combining FC modules, illustrating the orthogonality of higher-order statistics and FCs. Therefore, INSTA-BNN offers an additional choice to balance between accuracy and operations/parameters, and this extension capability is superior to DyBNN. Additionally, we provide hardware deployment results and optimization methods, which are critical directions for the practical use of BNN works.
\end{document}